# K-Splits: Improved K-Means Clustering Algorithm to Automatically Detect the Number of Clusters

*Seyed Omid Mohammadi[1], Ahmad Kalhor[2], Hossein Bodaghi[3]

University of Tehran, College of Engineering, School of Electrical and Computer Engineering, Tehran, Iran

[1]`S.OmidMohammadi@alumni.ut.ac.ir`, [2]`AKalhor@ut.ac.ir`,
[3]`Hossein.Bodaghi@ut.ac.ir`

\* Corresponding Author

**Abstract**. This paper introduces k-splits, an improved hierarchical algorithm based on k-means to cluster data without prior knowledge of the number of clusters. K-splits starts from a small number of clusters and uses the most significant data distribution axis to split these clusters incrementally into better fits if needed. Accuracy and speed are two main advantages of the proposed method. We experiment on six synthetic benchmark datasets plus two real-world datasets MNIST and Fashion-MNIST, to prove that our algorithm has excellent accuracy in automatically finding the correct number of clusters under different conditions. We also show that k-splits is faster than similar methods and can even be faster than the standard k-means in lower dimensions. Finally, we suggest using k-splits to uncover the exact position of centroids and then input them as initial points to the k-means algorithm to fine-tune the results.

**Keywords:** Data Clustering, Initialization, K-means Algorithm, Number of Clusters.

## 1      Introduction

Recent advances in scientific data collection technologies and the ever-growing volume of complex and diverse data make it harder for us to manipulate them or extract useful information. Moreover, most of this data is unlabeled, as finding suitable labels can be costly and time-consuming. Here is where unsupervised clustering methods come to assist. Clustering is the act of grouping items together which have similar characteristics and features. This way, each group (called a cluster) consists of similar items dissimilar to the other clusters' items.

K-means is a popular unsupervised clustering algorithm widely used due to its simple implementation and reasonably good results. However, this algorithm has its downsides, including high execution time and the dependency of final results on initial configurations. We also need to know the exact number of clusters (*k*) before proceeding with k-means, which can be a tricky task, especially for high-dimensional data. Setting a large *k* can cause many dead clusters (clusters with very few items), whereas a small *k* forces the items into insufficient clusters, leading to poor results. Multiple researchers tried to address this problem and introduced methods to estimate the number of clusters, including the Elbow method [1], the Silhouette method [2], and numerous variations of Subtractive clustering [3]. However, some of them require an exhaustive search or have a high computational cost.

This paper proposes a hierarchical algorithm wrapped around k-means to systemically and automatically determine the best number of clusters. The novelty of the proposed algorithm is that it uses the most significant data distribution axis to split the clusters incrementally into better fits if needed, causing a significant boost to the accuracy and speed. First, in Sec. 2, we present a short review of the k-means algorithm and discuss its limitations and related works. Then, in Sec. 3, we explain our proposed algorithm, and finally, we present experimental results in Sec. 4, followed by a conclusion.

## 2      Related Works: K-means Algorithm and Limitations

K-means is one of the widely used algorithms in data mining. Scalability and simplicity of implementation make the k-means algorithm a perfect candidate for practical applications ranging from optimization and signal processing to face detection [4].

First published in 1956 [5], k-means soon gained popularity using Lloyd's algorithm [6]. The algorithm starts by breaking the data into k clusters. As this initialization heavily impacts the final results, numerous researchers such as [7], [8], and [9] suggested different initialization schemes during the years. A straightforward method is Random initialization, which sets k random data points as cluster centers (called centroids). Next, the distances (usually Euclidean distance) between centroids and all data points are calculated, and each data point is assigned to the nearest cluster. The next step calculates the mean of all items in each cluster and sets new centroids. All data points are then checked against new centroids, and any relocation of data points to another cluster is done if needed. By repeating these steps, eventually, these once random centroids move step by step until we meet the convergence criterion, which is when all centroids are stable, and no change is needed. This process is shown in Algorithm 1.

K-means algorithm has several significant limitations, including:

1. The number of clusters (*k*): The standard k-means algorithm needs a predefined *k* to start, obtaining the exact value of which can be challenging in complex and multidimensional data. Multiple methods exist to estimate this setting, but the computational overhead is too much for complex and large data sets [10]. Wrapper methods like x-means and g-means also try to find the best value of $k$ by splitting or joining clusters. X-means uses the Bayesian Information Criterion penalty for model complexity [11], and g-means uses a statistical test to find Gaussian centers [12].
2. Centroids initialization: Solving this issue is very important because initialization seriously affects the final results. Reference [13] proposes a method of finding these centroids, which leads to better accuracy.
3. Time consumption: Calculating the distance between all data points and centroids repeatedly through iterations makes it a time-demanding process, especially for large amounts of data. Researchers usually try to tackle this problem through one of the following methods:

   - Implementing k-means on parallel platforms: These methods focus on implementing the algorithm on parallel GPUs or other distributed platforms to increase temporal efficiency [14][15].
   - Optimizing the algorithm: These methods try to enhance the algorithm to reduce computational complexity [16][17]. Even small reductions in the run-time can cause a big difference when dealing with large amounts of data.

**Algorithm 1:** The Standard K-Means Algorithm ($X, k$)

1. Choose $k$ data points as cluster centroids.

2. Calculate the distance of all data points from centroids and assign each data point to the nearest cluster.

3. Calculate the mean value of items in each cluster and recalculate new centroids of clusters.

4. If none of the centroids changed, proceed to step 5; otherwise, go back to step 2.

5. Output the results.

End of the algorithm.

## 3     Proposed Algorithm

This section proposes a new method of k-means based clustering algorithm called "k-splits" to systemically and automatically find the correct number of clusters (k) along the way. This hierarchical algorithm starts from a small number of clusters and splits them into more clusters if needed. The main advantage of k-splits, making it superior to others mentioned before, is that other methods split the clusters then run multiple tests to understand if it was the right decision, but we do the opposite.

As a result, the algorithm intelligently chooses the right cluster to break (we call it the worst cluster) and only focuses on one cluster at a time, which saves us much computational effort.

The main idea comes from the tendency of the k-means algorithm to find clusters of spherical shapes. Hopefully, we can separate the data into more relevant clusters by finding the axes of variance and data density in different areas. A perfect example is a cluster in a dumbbell shape. We can easily split this cluster by a hyperplane perpendicular to its most significant variance axis orientation.

The complete algorithm of the proposed k-splits is shown in Algorithm 2. The algorithm starts with one cluster (or more, based on prior knowledge), assuming all the data belong to one cluster, then it splits this massive cluster into two smaller clusters. It does this separation by finding the axis with the most significant variance. This axis has the same orientation as the eigenvector relevant to the data points' covariance matrix's maximum eigenvalue.

**Algorithm 2:** Proposed K-splits Algorithm $(X, \beta)$

1. Start with $k=1$.

2. Calculate $I^C$ for each cluster using (10) and $J_k$ using (9), then find the worst cluster ($C^w$) using (12).

3. Split the worst cluster into two clusters by assigning items to each cluster based on (5), then run Algorithm 1 (standard k-means) for these two clusters to obtain centroids $(c_{kw_1}, c_{kw_2})$.

4. If this is the first iteration; Use (8) to calculate the reference distance $d_{base}$ between these two centroids.

5. Else:

6.     Calculate centroid distances between every two clusters using (13) and set the minimum distance as $d$.

7. If condition (14) is not satisfied:

8.     Update $k \leftarrow k + 1$ and go to step 2.

9. Else:

10.     Discard centroids $(c_{kw_1}, c_{kw_2})$ and output the results of the iteration, which satisfies condition (15).

End of the algorithm.

For data $X^C$ in cluster $C$ with centroid $c$, we can calculate the covariance matrix $\Sigma^C$ from

$$\Sigma^C = \frac{1}{Q_C} \tilde{X}^{C\,T} \tilde{X}^C \quad (1)$$

where $Q_C$ is the number of items in cluster C, and

$$\tilde{X}^C = (X^C - c), c = \bar{X}^C \quad (2)$$

with $c$ being the centroid, $\bar{X}^C$ being the mean value of all items in each cluster $C$ and also

$$X^C = \begin{bmatrix} x_{C\,1} \\ \vdots \\ x_{C\,Q_C} \end{bmatrix} \quad (3)$$

Then we extract eigenvalues ($\lambda$) and eigenvectors ($v$) from this matrix. The method of accomplishing this goal can significantly impact the final computational complexity and run-time of this algorithm. However, for simplicity, one can use the SVD (Singular Value Decomposition) method [18] to obtain these values and then sort them in descending order.

$$\Sigma^C \xrightarrow{SVD} [\lambda_1^C \cdots \lambda_n^C], [v_1^C \cdots v_n^C] \qquad \lambda_1^C \gg \lambda_2^C \gg \cdots \gg \lambda_n^C \quad (4)$$

After that, we can find the hyperplane perpendicular to the most significant variance axis and assign data on different sides of it to two separate clusters:

$$\forall x^q: \quad \begin{cases} if\ \tilde{x}^q V^1 \geq 0 \ \rightarrow\ \tilde{x}^q \in \text{First cluster} \\ if\ \tilde{x}^q V^1 < 0 \ \rightarrow\ \tilde{x}^q \in \text{Second cluster} \end{cases} \quad (5)$$

where:

$$V^1 = \begin{bmatrix} v_1^1 \\ \vdots \\ v_n^1 \end{bmatrix}_{n \times 1} \quad (6)$$

and

$$\text{For } q = 1, \cdots, Q_C: \qquad x^q - c = \tilde{x}^q \quad (7)$$

This step assigns data points to these two clusters. We obtain an estimation of two initial centroids ($\tilde{c}_1, \tilde{c}_2$) by averaging data points in each cluster. Now to fine-tune the centers, we run the k-means algorithm as in Algorithm 1 on these two clusters until convergence to find the final centroids ($c_1, c_2$). Then we calculate the distance between these two centroids using (8). Any distance can be used here, but we proceed using $l_2$-norm.

$$d_{base} = \|c_1 - c_2\|_2 \quad (8)$$

The $d_{base}$ distance is the longest distance between two centroids, and we use it as a stop condition in future steps to determine when to end the process.

By now, the algorithm has formed only two clusters. From this point forward, in each iteration ($k$), the algorithm finds the worst cluster $C^w$, the best candidate for separation, then repeats all these mentioned steps and splits these worst clusters into two smaller ones. This process adds one cluster ($k \leftarrow k + 1$) with each iteration.

In each iteration, we also check the ratio of total items in each cluster to its covariance matrix's greatest eigenvalue and save the average value $J_k$ from (9) for later use. $J_k$ shows the density of clusters in each iteration and makes sure the algorithm is not over-splitting.

$$\text{For } C = 1, \cdots, k: \qquad J^C = \frac{Q_C}{\lambda_1^C} \quad , \quad J_k = \frac{\sum_{C=1}^{k} J^C}{k} \quad (9)$$

Testing the clusters, finding the worst cluster, and splitting it into two, increases the algorithm's time efficiency. Although computationally complex, running this test bypasses multiple unnecessary and exhaustive iterations of k-means, which leads to better run-time, especially on large data sets.

We introduce $I^C$, a criterion to help us determine the worst cluster (the cluster which needs further splitting). Choosing the worst cluster is an essential part of the process; that is why it takes finesse. The algorithm checks multiple elements to guarantee that we are splitting the right cluster. One of them is $\lambda_1^C$, giving us an idea of diversity and data distribution in that cluster. Another insight can come from *thr*, a threshold made of a combination of $Q, Q_C$, and $k$ as in (11), which shows the density of data in that cluster with a hint of the whole clustering situation the algorithm is in now, taking into account the within distance of each cluster. We combine these information sources to calculate $I^C$. However, one last detail remains to complete this process.

While splitting clusters, especially with unbalanced data, we may encounter massive clusters with slight variations that do not need separation. The algorithm might get stuck in splitting these clusters (we call Black Holes), given the data's unbalance. These Black Holes take much effort from the algorithm causing it to overlook other significant clusters and waste time and computational effort. We solve this issue by applying tanh to create a soft saturation as in (10).

$$I^C = \tanh(\frac{Q_C}{thr}) \lambda_1^C \qquad (10)$$

where:

$$thr = \frac{Q}{k} \quad , \quad \sum_{C=1}^{k} Q_C = Q \qquad (11)$$

Calculating $I^C$ from (10), helps determine the worst cluster. The cluster with the highest value of $I^C$ according to (12) is the candidate for further splitting.

$$C^w = \arg\max I^C \qquad (12)$$

After pinpointing the worst cluster, the algorithm repeats all previous splitting steps precisely, with one difference. This time after finding the centroids of sub-clusters ($c_{kw_1}, c_{kw_2}$), we calculate the distance of all centroids two by two and set the minimum distance as $d$ according to (13).

The final step in the algorithm is checking the stop condition (14). If the ratio of minimum cluster distance to maximum cluster distance is smaller than a threshold $\beta$, then no further separation is needed. This condition considers the extra-class distance of clusters, and $\beta$ can be set according to our prior knowledge of the data. Smaller $\beta$ values lead to more clusters, and larger values do the opposite. For very dense data, larger values of $\beta$ are suggested, and for more sparse data, vice versa.

For $i = 1, \cdots, k+1$, $j = 1, \cdots, k$, $i \neq j$:

$$d_{ij} = \|c_i - c_j\|_2 \quad , \quad d = \min_{\substack{i=1,\cdots,k+1 \\ j=1,\cdots,k \\ i \neq j}} d_{ij} \qquad (13)$$

$$\frac{d}{d_{base}} \leq \beta, \quad 0 < \beta < 1 \qquad (14)$$

After reaching the stop-condition, we can use the optional step (15) to choose the best iteration and discard later results to reduce redundant clusters and reduce sensitivity to hyperparameter $\beta$, which controls the separation condition, making the algorithm easier to use. However, experiments show that it is best to skip this step to get better results for very dense or highly overlapped (> 50%) data. So it is always good to run some tests on the density of data before starting the algorithm.

$$iter = \arg\max J_k \quad (15)$$

K-splits efficiently and automatically finds the number of clusters and their borders. However, the accuracy of assigning items to clusters might not be the highest because our primary focus is on finding centroids in the shortest time possible. This problem is solved via an optional fine-tuning step. Thus, it is advisable to use final centroids from k-splits and then run the conventional k-means on the data using these known points as initialization. This simple step can fine-tune k-splits and highly improve the final results, as shown in Table 2. and Fig. 1.

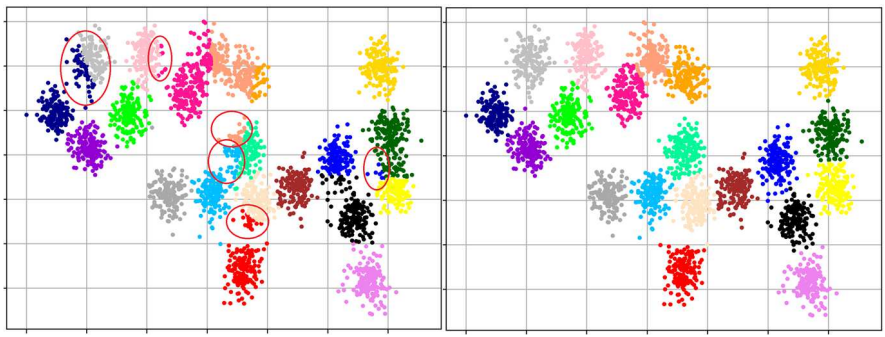

**Fig. 1.** Clusters before and after fine-tuning results of k-splits on the A1 dataset. The first graph highlights some incorrectly labeled data points and fine-tuning corrects them, as shown in the second graph.

## 4   Experiments

### 4.1   Accuracy and Run-time Comparison

For experimental analysis, we chose synthetic benchmark datasets carefully to show different challenging aspects of the data. Detailed information about the datasets can be found in [19]. Each dataset has unique properties, including varying cluster size (A), dimension (Dim), overlap (S), structure (Birch), balance (Unbalance), and a combination of dimension and overlap (G). We also use real-world datasets MNIST [20] and Fashion-MNIST [21] to further demonstrate the applicability of our method. Table 1. presents specifications for each dataset.

Although many advanced methods exist to find the correct number of clusters, our focus is only on k-means based algorithms. Thus, we compare our algorithm and its fine-tuned version with g-means and x-means, which are the most similar and comparable methods to k-splits. We also use k-means (with the correct number of $k$ as input) as our baseline. A tricky part of the standard k-means is knowing the exact number of clusters. Here we input the correct number of $k$ to test k-means, so we expect much better results. K-means is highly affected by initialization, and these wrapper methods might find local optimum, so it is known to run these algorithms multiple times and report the best results. Therefore, it is customary to report the results of a ten repeat (10R) of each of these three algorithms. Although comparing the run-time of a 10R algorithm with one repeat of our proposed algorithm might seem unfair, bear in mind that most of the time, a single run of those other algorithms does not lead to acceptable results. In contrast, k-splits needs no repetition, leading to deterministic results that remain unchanged through reruns.

Experiments are conducted using Python 3.7.4 on a system with Intel Core i5-4200M CPU@ 2.50 GHz, 4GB memory, and 1T hard disk space. We use Scikit-learn package implementation of k-means and Pyclustering package implementation of g-means and x-means. We also use Numpy.linalg, which is implemented using LAPACK routines, to obtain eigenvalues and eigenvectors. We set $\beta = 0.01$ for half the experiments, and for the other half containing medium density and overlap such as A and S, $\beta = 0.1$ was used. For pictorial datasets MNIST and Fashion-MNIST, we first use PCA (Principal Component Analysis) with 0.9 variance to reduce the original dimension size of 784 to 87 and 84 respectively, then set $\beta = 0.95$ due to the density of data points and distances and apply the algorithm.

**Table 1.** Datasets Specifications.

| Dataset | N | C | D | Overlap | Dataset | N | C | D | Overlap |
|---|---|---|---|---|---|---|---|---|---|
| *A1* | 3000 | 20 | 2 | 20% | *G2-2-30* | 2048 | 2 | 2 | 15% |
| *A2* | 5250 | 35 | 2 | 20% | *G2-2-50* | 2048 | 2 | 2 | 43% |
| *A3* | 7500 | 50 | 2 | 20% | *G2-128-10* | 2048 | 2 | 128 | 13% |
| *S1* | 5000 | 15 | 2 | 9% | *Birch1* | 100k | 100 | 2 | 52% |
| *S2* | 5000 | 15 | 2 | 22% | *Birch2* | 100k | 100 | 2 | 4% |
| *Dim32* | 1024 | 16 | 32 | 0% | *MNIST* | 60k | 10 | 87 | - |
| *Dim1024* | 1024 | 16 | 1024 | 0% | *F-MNIST* | 60k | 10 | 84 | - |
| *Unbalance* | 6500 | 8 | 2 | 0% | | | | | |

N: Number of Data Points, C: Number of Clusters, D: Dimensions

The results are summarized in Table 2. Each value is the mean across five validation folds. We report the number of detected clusters ($k$), execution time (t reported in seconds), and an external validity measure called the *adjusted rand index* (ARI). ARI is a score that shows the similarity between two sets of clustering results and is equal to 0 for random results and 1 for exact clustering matches. For each dataset, the best instances of the number of predicted clusters ($k$), execution time (t), and ARI are in bold font with some exceptions. One exception is that we can only compare execution time if both methods provide "acceptable" results; thus, lower run-time with terrible results in Table 2 are only underlined and not bolded. The other exception is that fine-tuned k-splits is the improved version of k-splits; thus, it is enough to compare the fine-tuned version's execution time.

**Table 2.** Comparison Results of the Algorithms.

| Dataset | K-splits | | | Fine-tuned K-splits | | 10R G-means | | | 10R X-means | | | 10R K-means | |
|---|---|---|---|---|---|---|---|---|---|---|---|---|---|
| | k | t (s) | ARI | t (s) | ARI | k | t (s) | ARI | k | t (s) | ARI | t (s) | ARI |
| *A1* | **20** | 0.10 | 0.80 | **0.14** | **1.0** | 27 | 0.17 | 0.89 | **20** | 0.16 | **1.0** | 0.18 | **1.0** |
| *A2* | 35 | 0.22 | 0.87 | **0.23** | **1.0** | 41.2 | 0.37 | 0.94 | 20 | 0.27 | 0.62 | 0.54 | 0.99 |
| *A3* | **50** | 0.33 | 0.90 | **0.38** | 0.97 | 57.8 | 0.86 | 0.95 | 4 | <u>0.18</u> | 0.11 | 0.86 | **0.99** |
| *S1* | **16** | 0.27 | 0.96 | 0.27 | 0.98 | 233.4 | 1.73 | 0.19 | 20 | 0.40 | 0.94 | **0.14** | **0.99** |
| *S2* | **16** | 0.29 | 0.87 | 0.31 | 0.92 | 194.2 | 1.93 | 0.20 | 18.6 | 0.37 | 0.90 | **0.20** | **0.94** |
| *Dim32* | **16** | 0.13 | **1.0** | 0.13 | **1.0** | 649.6 | 1.42 | 0.07 | 16.8 | **0.07** | **1.0** | 0.08 | **1.0** |
| *Dim1024* | 17 | 310 | **1.0** | 313 | **1.0** | 827.4 | 36.3 | 0.03 | **16** | 0.91 | **1.0** | 0.79 | **1.0** |
| *G2-2-30* | **2** | 0.01 | **0.96** | **0.01** | **0.96** | **2** | 0.06 | 0.95 | **2** | 0.06 | 0.95 | 0.02 | **0.96** |
| *G2-2-50* | **2** | 0.01 | **0.70** | **0.01** | **0.70** | **2** | 0.06 | **0.70** | **2** | 0.04 | **0.70** | 0.03 | **0.70** |
| *G2-128-10* | **2** | 0.07 | **1.0** | 0.09 | **1.0** | 779 | 6.60 | 0.00 | **2** | 0.24 | **1.0** | 0.11 | **1.0** |
| *Birch1* | 10 | 2.36 | 0.66 | **4.85** | 0.94 | 2562 | 741 | 0.08 | 4 | <u>3.66</u> | 0.06 | 48.32 | **0.95** |
| *Birch2* | **101** | 2.04 | 0.98 | **2.58** | **1.0** | 126.6 | 8.51 | 0.93 | 20 | 4.77 | 0.30 | 15.44 | **1.0** |
| *Unbalance* | **10** | 0.28 | **1.0** | 0.30 | **1.0** | 14.2 | 0.46 | **1.0** | 4 | 0.24 | 0.99 | **0.08** | **1.0** |
| *MNIST* | 19 | 3.80 | 0.20 | **7.16** | 0.36 | 11.9k | 16h | 0.00 | 20 | 71.6 | **0.36** | 25.54 | **0.36** |
| *F-MNIST* | 7 | 1.96 | 0.21 | **2.98** | 0.37 | 14.2k | 21h | 0.00 | 20 | 57.5 | 0.34 | 15.63 | 0.35 |

k: Number of predicted clusters, t (s): Execution time in seconds

In predicting the number of clusters, k-splits' predictions are the closest to reality, except for Dim1024, and even there, the error is minimal. Whereas g-means almost always over-splits the data, with many unacceptable results. One extreme example is the MNIST and F-MNIST datasets, for which the algorithm predicts more than ten thousand clusters! This might be an excellent example of the black hole problem mentioned before; getting stuck in very dense clusters misguides the algorithm, leads to false results, and wastes time and effort, the exact thing we avoid by implementing multiple stop-conditions in k-splits. X-means, on the other hand, seems to under-split in many cases. K-splits accurately predicts the number of clusters in almost all cases. Some examples of clustering results using k-splits are shown in Fig. 2.

Regarding the execution time, g-means is always slower, especially for large (and dense) datasets; the black hole problem completely ruins the process leading to extreme examples like the Fashion-MNIST dataset, which took 21 hours to be processed. The x-means algorithm takes an acceptable amount of time to finish, but we should consider that under-splitting of data might be a factor in that. K-splits (even the fine-tuned version) is faster than g-means and x-means under almost all situations, with few exceptions. It is also faster than the conventional k-means in many cases.

The time efficiency of k-splits is the result of three decisions in each iteration:

- It only splits the so-called "the worst cluster" and keeps other clusters intact.
- It only uses the k-means algorithm to separate only two clusters in each step, which is very simple to solve.
- It gives the k-means algorithm accurately calculated initial centers leading to much faster convergence and better results than random initialization.

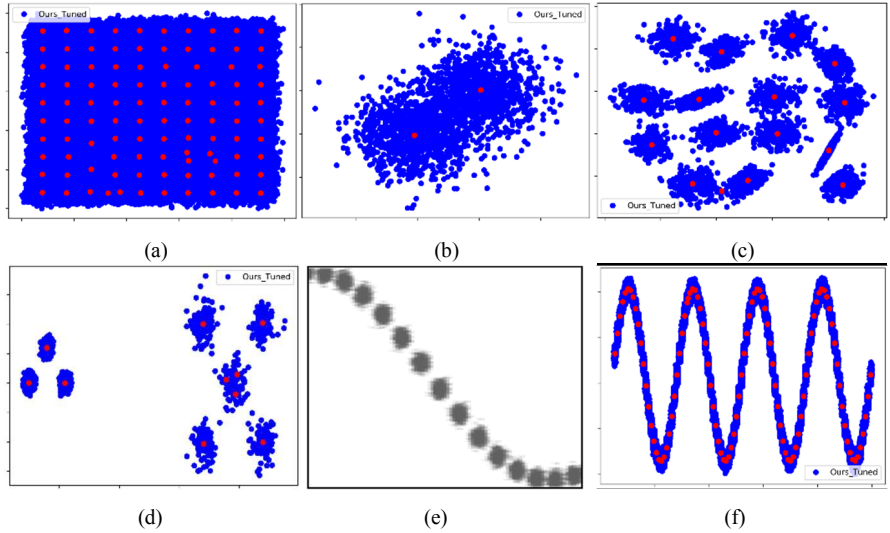

**Fig. 2.** Predicted centroids using k-splits on benchmark datasets. (a) Birch, (b) G2-2-50, (c) S1 and (d) Unbalance datasets. (e) and (f) Zoomed structure of Birch2 and our results.

High dimensionality harms the time efficiency of the k-splits algorithm. The type of computations we use grows in difficulty and negates the aforementioned positive effects in higher dimensions. Although the algorithm yields good results, the time consumption in dimensions higher than 1000 is high. Therefore, a dimension reduction before using this kind of data can be helpful.

Comparing the ARI values in Table 2. proves that k-splits is always more accurate than both g-means and x-means in assigning each data point to the correct cluster. We can also see that it is always a good idea to fine-tune k-splits which leads to much better results, some of them even better than the standard k-means. That is because k-splits pinpoints the centroids, which is an excellent boost for the k-means algorithm. The time cost of fine-tuning is also acceptable. In some cases, like A1 and A2 datasets, the fine-tuned k-splits is still more time-efficient than the standard k-means, and in some unique structures like Birch, this time gap is more significant.

We should bear in mind that the main superiority of k-splits is the ability to find the correct number of clusters automatically. Hence, we do not expect more accurate or faster results than the original k-means algorithm with the right $k$ as input.

However, as shown in Table 2., the execution time of k-splits is faster than k-means in some cases, and the performance is acceptable. Part of this performance gap is due to the rigidness of k-splits. Once a cluster is separated into two, those data points cannot move to other clusters. The combination of k-splits and k-means, introduced as fine-tuned k-splits, solves this problem. Fine-tuned k-splits benefits from the speed and accuracy of k-splits in finding the right $k$ and the high performance of standard k-means in assigning each data point to the clusters.

### 4.2  Effect of Hyperparameter $\beta$

We experiment with different values of hyperparameter $\beta$ on a synthetic dataset to investigate the sensitivity of the algorithm to this hyperparameter. We conduct two tests, one without applying condition (15) and another with this condition and using $J_K$ as a second stop criterion, the results are shown in Fig. 3.

The synthetic dataset used in these experiments has a size of $N = 10{,}000$ data points distributed in ten clusters and ten dimensions. It is evident from Fig. 3. (a) that k-splits successfully predicts the correct number of clusters in a wide range of $\beta$, but outside this range, the algorithm fails. However, Fig. 3. (b) shows that using $J_K$ as a secondary stop condition helps decrease this sensitivity and provides acceptable results for almost the whole range. Nonetheless, the problem with $\beta$ is not entirely solved. Firstly, as mentioned in Sec. 3, $J_K$ cannot be used for highly overlapped and dense data, and secondly, although $\beta$ has an acceptable range for each dataset, this range changes as the structure of the dataset, especially the density, changes.

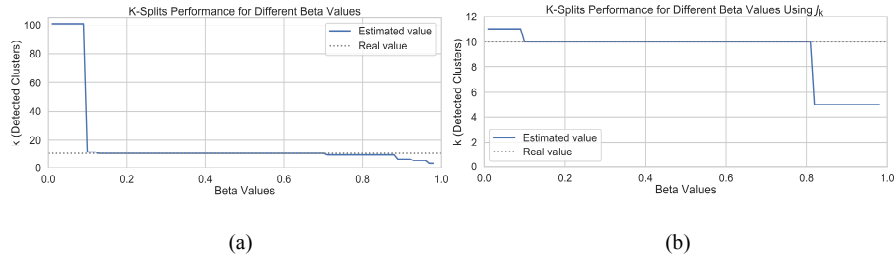

(a)          (b)

**Fig. 3.** K-Splits performance in finding ten clusters: (a) Using only β as stop condition, (b) Using $J_K$ as stop condition.

### 4.3  Effects of Dataset Parameters

We also conduct several other experiments to analyze further the effects of dataset parameters on k-splits. The datasets used in this section share the specifications mentioned above, and in each experiment, only one parameter of the dataset is analyzed. These parameters are dataset size, number of clusters, and dimension size. Changing the parameters leads to a different set of data with a new structure; thus, we utilize a fivefold scheme to calculate each point and create regression plots using central tendency and confidence intervals.

**Dataset Size (N).** For this experiment, dataset sizes in a range of [500, 30000] are analyzed. This test provides a good sense of the time complexity of methods; that is why we conduct it on k-splits and 10R versions of k-means, x-means, and g-mean simultaneously. The results are shown in Fig. 4.

As mentioned earlier, the g-means algorithm experiences critical problems, getting stuck in local optimum and black holes. These problems are readily detectable in Fig. 4. (a) as in some points, we witness abnormalities in the form of strong peaks in execution time. In this example, these peaks get as high as 800 seconds, while execution time for other methods and even other points of the same algorithm is well below 5 seconds. In conclusion, g-means is the slowest method among these methods.

To further compare these methods, in the next test, we assume that g-means never encounters the black hole problem, and we discard all these problematic points; the result is Fig. 4. (b). We can see that regardless of the problems caused by getting stuck, g-means is still the slowest algorithm. After that, we have x-means and k-means, respectively, and both k-splits and its fine-tuned version execute faster than all three methods.

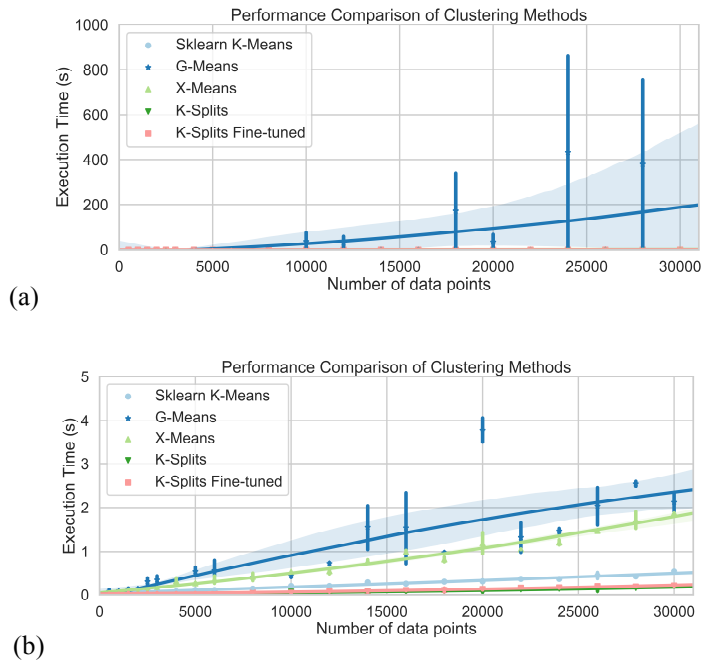

**Fig. 4.** Comparison of methods regarding execution time. (a) Actual values (b) Values in case g-means never encounters the black hole problem.

**Number of Clusters (C).** The number of clusters is another critical parameter in each dataset. Therefore, we test the k-splits algorithm under different numbers of clusters in the range of [2, 100]. The results are presented in Fig. 5. Each dataset instance has a different structure; moreover, changing the number of clusters in a limited space changes the density and overlap. Thus, as presented in Fig. 5. (a), the algorithm cannot accurately predict larger cluster numbers using one constant $\beta = 0.5$. However, if we schedule $\beta$ to change through the experiment based on our prior knowledge of the structure (similar to what we do in real applications), the performance increases according to Fig. 5. (b).

Needless to point out that although we can achieve great results using a good guess about the range of $\beta$, it is still one of the main downsides of the k-splits algorithm. Different cluster sizes require a different range of $\beta$, a good guess of which is needed to predict the number of clusters in turn! That is why checking the density of data is a good step in setting hyperparameter $\beta$. We analyze the effect of the number of clusters on execution time, and the results are of order two and presented in Fig. 5. (c).

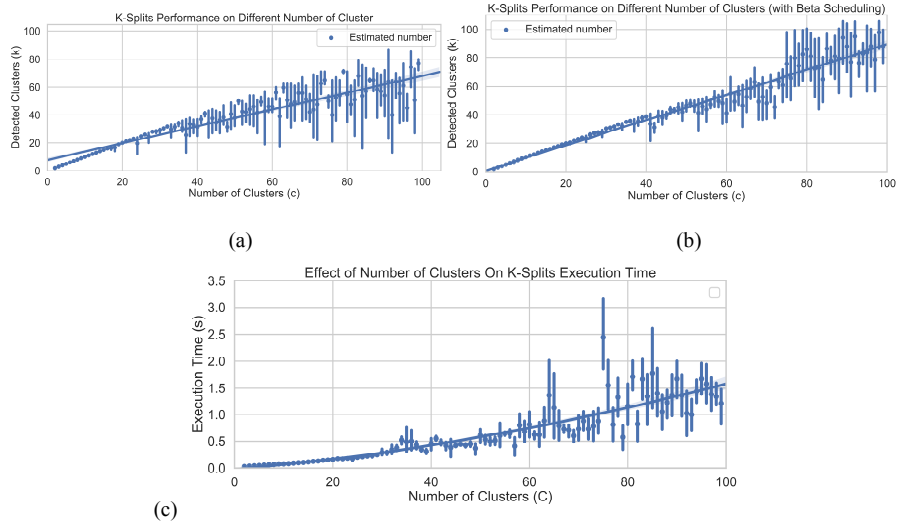

**Fig. 5.** Effect of the number of clusters on k-splits. (a) Performance using a constant β=0.5. (b) Performance using a variable β. (c) Effect of the number of clusters on execution time.

**Data Dimension Size (D).** As our last experiment, we check the effect of dataset dimension on k-splits. We use a constant $\beta = 0.5$ throughout the test. The effects of dimension size on our proposed algorithm's accuracy and execution time are shown in Fig. 6. We can see that k-splits accurately estimates the number of clusters in different dimensions, except in about ten first instances, which have lower accuracy. This low accuracy might occur because of the data's higher density due to very low dimensions. Fig. 6. shows that dimension has minimal effect on the range of suitable $\beta$, but should not be neglected.

## 5  Conclusion

We have proposed k-splits: an incremental k-means based clustering algorithm to detect the number of clusters and centroids automatically. We also introduced a fine-tuned version of k-splits that uses these centroids as initialization for the standard k-means and dramatically improves the performance. We used six synthetic datasets and also MNIST and Fashion-MNIST datasets to show that k-splits can accurately find the correct number of clusters and pinpoint each cluster's center under different circumstances. K-splits is faster than similar methods like g-means and x-means and, in some cases, even faster than the standard k-means. The accuracy of the results and the performance are also higher than these methods. Furthermore, k-splits needs no repetition as it leads to deterministic results.

K-splits starts from a small number of clusters and further splits each cluster if needed. The starting point does not have to be one cluster and, if known, can be set by the user leading to an even faster result. The algorithm takes one threshold parameter $\beta$, which controls separation condition, as input which should be set based on our prior knowledge of the data density. Clustering in very high dimensional spaces with k-splits is time-consuming, so it is preferable to use dimension reduction techniques before applying the algorithm. We intend to focus on two improvements in future works. First, to optimize the calculations needed to obtain similar results, especially in higher dimensions, and second, to eliminate the need for any extra hyperparameters if possible.

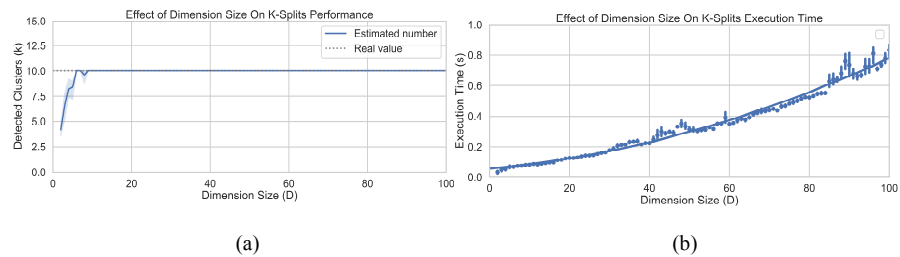

(a)  (b)

**Fig. 6.** The effect of data dimension on k-splits: (a) Effect on cluster prediction accuracy, (b) Effect on execution time.